\documentclass[runningheads]{llncs}
\usepackage{graphicx}
\usepackage{comment}
\usepackage{multirow}
\usepackage{amsmath,amssymb} %
\usepackage{color}
\usepackage{subcaption}
\usepackage{amsmath}
\usepackage{hyperref}
\usepackage{amssymb}
\usepackage{booktabs}
\usepackage[dvipsnames]{xcolor,colortbl}

\usepackage[symbol]{footmisc}

\begin{document}

\newcommand{\todo}[1]{\textcolor{Red}{\textbf{TODO: #1}}}
\newcommand{\aude}[1]{\textcolor{Red}{\textbf{ #1}}}
\newcommand{\alex}[1]{\textcolor{Green}{\textbf{Alex: #1}}}
\frenchspacing
\everymath{\displaystyle}

\pagestyle{headings}
\mainmatter

\title{We Have So Much In Common: Modeling Semantic Relational Set Abstractions in Videos}

\titlerunning{Semantic Relational Set Abstraction for Event Understanding}
\author{Alex Andonian* \inst{1} \and
Camilo Fosco* \inst{1} \and
Mathew Monfort\inst{1} \and  \\
Allen Lee\inst{1} \and 
Rogerio Feris\inst{2} \and
Carl Vondrick\inst{3} \and
Aude Oliva \inst{1} 
}

\authorrunning{A. Andonian et al.}
\institute{Massachusetts Institute of Technology \and MIT-IBM Watson AI Lab \and Columbia University}
\maketitle

\renewcommand{\thefootnote}{\fnsymbol{footnote}}
\footnotetext[1]{Equal contribution.}
\renewcommand{\thefootnote}{\arabic{footnote}}

\renewcommand{\thefootnote}{\fnsymbol{footnote}}

\begin{abstract}
Identifying common patterns among events is a key ability in human and machine perception, as it underlies intelligent decision making. We propose an approach for learning \emph{semantic relational set abstractions} on videos, inspired by human learning. We combine visual features with natural language supervision to generate high-level representations of similarities across a set of videos. This allows our model to perform cognitive tasks such as \emph{set abstraction} (which general concept is in common among a set of videos?), \emph{set completion} (which new video goes well with the set?), and \emph{odd one out detection} (which video does not belong to the set?). Experiments on two video benchmarks, Kinetics and Multi-Moments in Time, show that robust and versatile representations emerge when learning to recognize commonalities among sets. We compare our model to several baseline algorithms and show that significant improvements result from explicitly learning relational abstractions with semantic supervision. Code and models are available online
\footnote[2]{Project website: \url{abstraction.csail.mit.edu}}.

\end{abstract}

\section{Introduction}

\begin{figure}[t]
    \centering
    \includegraphics[width=0.8\linewidth]{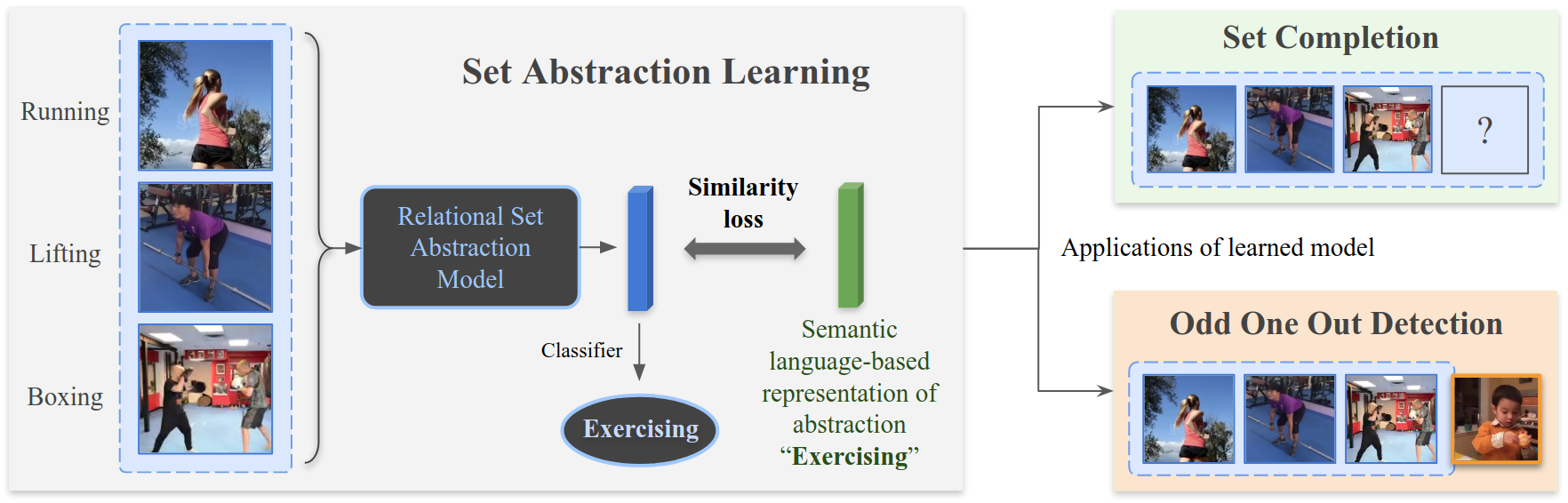}
    \caption{\textbf{Semantic Relational Set Abstraction and its Applications:} We propose a paradigm to learn the commonalities between events in a set (the \emph{set abstraction}) using a relational video model. Our model is trained to approximate the semantic language-based representation of the abstraction and predict the abstract class shared by the videos. Once trained, the model is able to identify the abstraction that represents a set of videos, select videos that fit this abstraction, and detect when a member of the set does not match the common theme.}
    \label{fig:teaser}
\end{figure}

Humans are extraordinary at picking out patterns between different events, detecting what they have in common and organizing them into abstract categories, a key reasoning ability. Our goal in this paper is to instantiate this ability into a computer vision system. Learning the semantic relational abstraction between a set of events (Figure \ref{fig:teaser}) allows a model to perform cognitive-level tasks similar to a person abstracting common patterns. If a model has learned that \emph{exercising} can take many forms (\emph{running}, \emph{weightlifting}, \emph{boxing}), its feature representation can naturally be used to select which new event is similar to the set, or detect an incompatible exemplar. Enriching the learning of these abstractions with semantic and verbal content, allows us to move one step closer to people's ability to combine visual and contextual relationships for deeper event understanding.

In this paper, we propose an approach for learning \emph{semantic relational set abstraction} which recasts a single exemplar recognition task to the task of encoding conceptual relationships shared among a set of videos.  We apply our trained model to solve a variety of operations, namely \emph{set abstraction} (what is in common?), \emph{set completion} (which new video goes well with the set?), and \emph{odd one out detection} (which video does not belong to the set?).  Additionally, we compare our abstraction model to human performance on a novel \emph{relational event abstraction} task where participants rank a set of query videos according to how closely they align with the abstract semantic relationship found between a set of reference videos.  This human baseline provides a strong evaluation metric for determining how well our model can encode semantic relationships and allows us to measure the human ability to abstract common concepts from video sets.

By formulating the new set abstraction task in tandem with a language supervision module, we aim to better approximate human cognitive-level decisions. Importantly, we pioneer a data generation methodology which approximates human behavior in abstraction-related tasks. The semantic relational algorithm allows us to sample training examples that categorize the commonalities between sets of videos in a human understandable way. We root our experiments in event understanding, leveraging the large scale video datasets Kinetics \cite{kinetics} and Multi-Moments in Time \cite{monfort2019multimoments}, replicating all our results with these two benchmarks.
To summarize, the main contributions of this paper are:
\begin{enumerate} 
\item  A novel \textbf{relational set abstraction model} which generates representations of \emph{abstract events} relating a set of videos in a language-based geometric space and assigns a human-understandable label to the common concept underlying each set. This can be used for cognitive-level relation tasks, namely \emph{set completion} and \emph{odd one out detection}, achieving human-level performance. 
\item A novel paradigm, the \textbf{Relational Event Abstraction task}, which measures human performance on event abstraction. Given a set of reference videos representing specific events (e.g. \emph{digging}, \emph{peeling}, \emph{unwrapping}), the task involves finding the common abstract concept shared by the videos in the set (e.g. \emph{removing} in this case), and ranking a set of query videos based on how close they align with the abstraction.
\item A \textbf{dataset for Relational Event Abstraction} built using a novel \emph{semantic relational algorithmic methodology} rooted in natural language which correlates highly with human results on the \emph{relational event abstraction} task. This allows us to sample a large number of reference and query sets for training and evaluation, avoiding expensive human annotation. 
\end{enumerate}

\section{Related Work}

\noindent \textbf{Concepts Organization.} Concepts can be organized in a hierarchical structure (i.e. trees), a chain (i.e. linear organization), or a ring (i.e. perceptual similarities of colors)\cite{tenenbaum2011}. Computer vision work on visual classification most often uses trees, with root categories forming the base of taxonomies (i.e. for object  \cite{deng2014large} and scene classes \cite{xiao2016sun}). Hierarchies can be pre-defined  \cite{jia2013visual,verma2012learning} or learned \cite{bannour2012hierarchical,deng2011fast}, and can help with transfer learning between categories \cite{lim2011transfer}, and class prediction for videos \cite{nauata2018structured,nauata2017hierarchical}. EventNet \cite{Ye:2015:ELS:2733373.2806221} built an action concept dataset with a hierarchical structure that includes low-level event labels as leaf nodes and increasingly abstract concept labels as parent nodes.  They trained a CNN to identify the low-level event labels from the video frames and combine the representation learned from this model with a set of SVMs to predict the higher-level concepts associated with the video.  Here, we similarly use a pre-defined relational organization between activities (i.e. \emph{jog}, \emph{swim} and \emph{weightlift} all share the abstract relation \emph{exercise}) as a tool for learning the abstract semantic relations between sets of videos. While previous works consider a single instance at a time, our goal is to generate representations for \emph{sets} of videos to identify common relationships. \\
\noindent \textbf{Video Recognition.} 
Two stream convolutional networks (CNNs) \cite{simonyan2014two} combine static images and optical flow. In \cite{donahue2015long}, a recurrent model uses an LSTM to learn temporal relationships between features extracted from each frame.  3D CNNs \cite{tran2015learning} aim to directly learn motion using 3D convolutional kernels to extract features from a dense sequence of frames.  I3D proposes incorporating optical flow with 3D CNNs to form a two stream 3D network \cite{carreira2017quo} ``inflated'' from 2D filters pre-trained on ImageNet \cite{deng2009imagenet}.  Temporal Segment \cite{wang2016temporal} and Temporal Relation Networks \cite{zhou2017temporal} model relationships between frames from different time segments while non-local modules \cite{Wang_2018_CVPR} capture long-range dependencies.  SlowFast Networks \cite{Feichtenhofer_2019_ICCV} combine two streams using dense (fast) and sparse (slow) frame rates to simultaneously learn spatial and temporal information. \\
\noindent \textbf{Visual Similarity.}  Prior work has proposed methods for estimating similarity between images and video pairs for retrieval and anomaly detection. Fractal representations have been used to estimate pair-wise image similarity and relationships in order to solve the Odd One Out problem by encoding the spatial transformations needed to convert each image in a set to each other image \cite{McGreggor:2011:FOO:2069618.2069666}.  Semantic word similarity has been shown to be a good approximation for quantifying visual relationships \cite{4587822} while Siamese Networks \cite{Bromley:1993:SVU:2987189.2987282} have been used to naturally rank the similarity of sets of images for one-shot image recognition \cite{koch2015siamese}.  ViSiL \cite{Kordopatis-Zilos_2019_ICCV} proposes an architecture for learning the similarity between pairs of videos that incorporates both spatial and temporal similarity for video retrieval. Odd-one-out networks \cite{Fernando_2017_CVPR} learn temporal representations of videos that are used to identify the video in a set that has \emph{out of order} frames.  IECO \cite{Zhao_2019_ICCV} uses ideas from SlowFast Networks \cite{Feichtenhofer_2019_ICCV} to form a two-stream ECO network \cite{Zolfaghari_2018_ECCV} to learn instance, pose and action representations to estimate video similarity in retrieval.\\
\noindent \textbf{Learning Semantic Relationships in Visual Data.} Different approaches have been proposed for learning semantic relationships between sets of images and videos.  Reinforcement learning is used as a method for selecting subsets of data that preserve abstract relationships \cite{Muhammad_2019_ICCV}.  A Bayesian model has been used with a conceptual hierarchy formed from ImageNet \cite{deng2009imagenet} to identify the concept relation in image sets \cite{NIPS2013_5205}.  %
Relation Networks have been used to infer object relations \cite{NIPS2017_7082} and Interaction Networks have helped to identify physical relations in complex systems \cite{NIPS2016_6418}.  
We extend the idea of relation learning to videos for learning event relations in sets of varying length rather than object relations between images pairs
Laso \cite{Alfassy_2019_CVPR} utilizes similarities, and differences, in object labels of images pairs to improve few shot learning by learning the union, intersection and subtraction between the binary label vectors for each image.  

We take a similar approach to learning the intersection operation of the ``abstract'' labels for sets of videos but we differ in that we operate on sets of varying size and are learning the common semantic abstractions of events shared in the set rather than the common labels found in video pairs. The most similar prior work ranks a set of unseen videos based on their similarity to a provided event description \cite{Chang:2015:SCD:2832415.2832559} by measuring the semantic correlation between the event and each individual concept in a dictionary of concepts (e.g bike, mountain, etc) with cosine distance on the word embedding generated from a skip-gram model \cite{NIPS2013_5021} trained on a large text corpus. A set of previously unseen videos is ranked according to the correlation between the video and the provided event using the event-concept correlation and the concept classifier for each video.  We utilize word embeddings to capture class relationships as was done in other methods that use similar embeddings to identify semantic visual relations \cite{Jain_2015_ICCV,DBLP:journals/corr/abs-1708-04014,4587822}.  However, we introduce a relational graph and abstract relational embeddings that are compounded by the embeddings of related classes (see Section \ref{sec:dataset}) and apply our approach to the task of recognizing the abstraction between a set of videos.

\section{A Dataset for Relational Event Abstraction}
\label{sec:dataset}
Our goal is to categorize the relationships between sets of videos close to human reasoning.  We build a dataset for identifying semantic relational event abstractions between sets using word embeddings from natural language \cite{fasttext} and from an abstraction graph where each node represents a semantic relation between its children. Word embeddings, which capture context and word-to-word relationships \cite{NIPS2013_5021} from a large text corpus, are complementary to our semantic abstraction graph and allow our model to capture relationships not directly encoded into the graph structure. Next, we provide an overview of our approach. %

\noindent \textbf{Video Datasets.} As an initial step for representing semantic set relationships, we focus on the \emph{relationships between activities in video clips}, using Kinetics \cite{kinetics} and Multi-Moments in Time (M-MiT) datasets \cite{monfort2019multimoments}.  The labels in Kinetics define specific event classes such as \emph{biking through snow} and \emph{cooking on campfire}. The labels in M-MiT are more general such as \emph{bicycling} and \emph{cooking}.  The contrast between the class structures allows us to validate our approach to settings with both low-level event categories (Kinetics) and high-level classes (M-MiT).

\begin{figure}[t]
\centering
\includegraphics[width=0.5\linewidth]{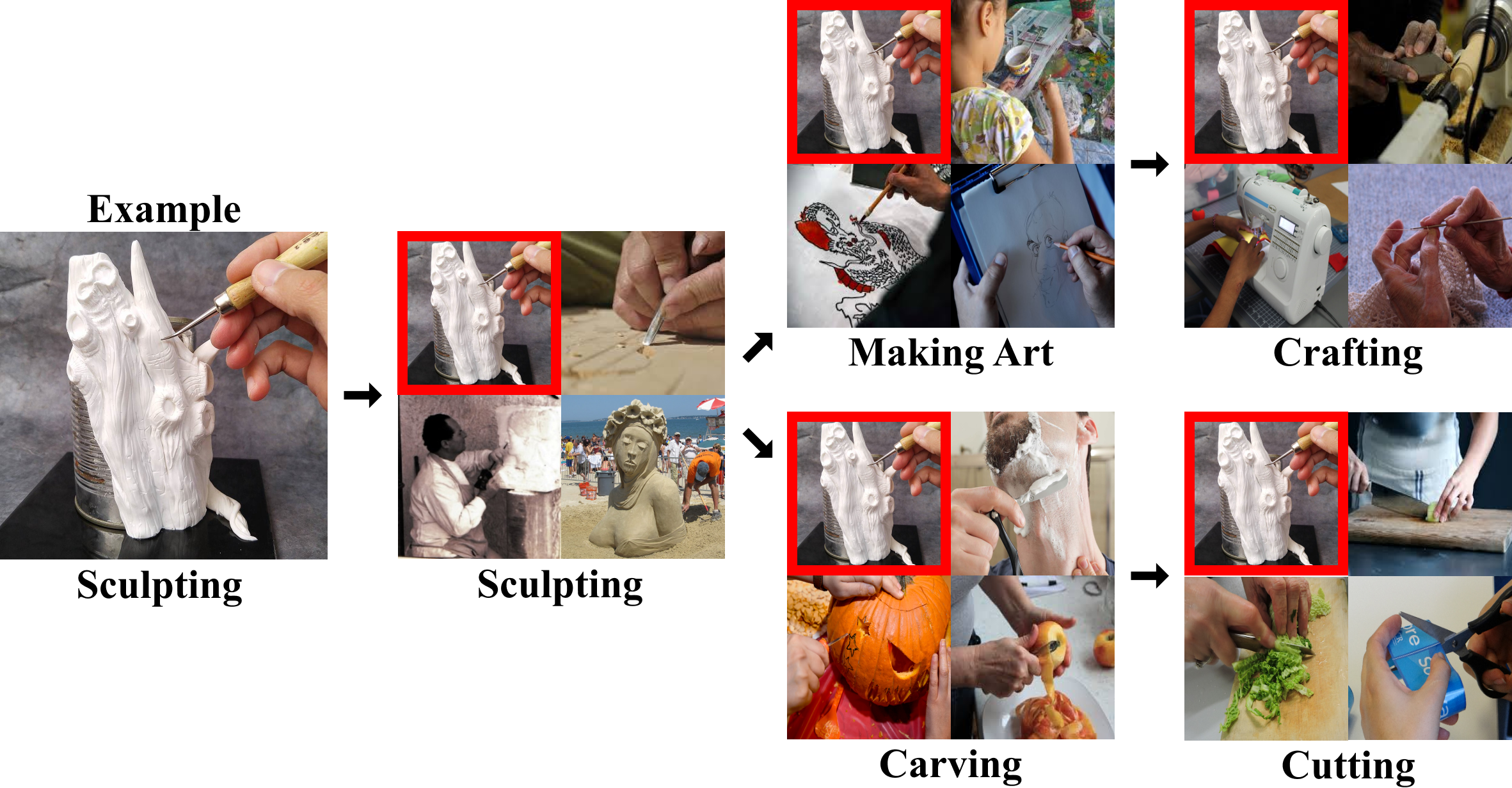}
\caption{\textbf{Semantic Relational Graph}: Example of a class relation in our semantic graph for M-MiT.  A specific video of \emph{a person sculpting} is an exemplar of the category \emph{sculpting}. The set of videos that share the activity \emph{sculpting} is itself a member of multiple sets that capture higher-level abstractions, like \emph{making art} and \emph{carving} which in turn are members of \emph{crafting} and \emph{cutting} respectively.}
\label{fig:semanticGraph}
\end{figure}

\noindent \textbf{Semantic Relational Graph.} To form our relational graphs, we start with the activity categories provided by the class vocabularies of Kinetics and M-MiT.  We assign each category to a synset in the WordNet lexical database \cite{miller1990introduction} that captures the specific meaning of the class label applied to its member videos.  Then we extract the hypernym paths of each class in WordNet and add the path, and the members of each path, to our graph.  We verify the graph by hand and add any missing relations not captured by the WordNet paths. Building the graph in this way for the full class vocabulary of each dataset allows us to form a trace from each low-level action to high-level categories that capture the abstract relationships between their descendant classes.  For example, in the Kinetics graph, the abstraction node for \emph{baking} has two children that share the relation of \emph{baking}, \emph{making a cake} and \emph{baking cookies}, and is a child of the abstraction node \emph{cooking} together with other categories such as \emph{frying} and \emph{cooking chicken} that share the same relation of \emph{cooking}.  We do not restrict the nodes in the graph to have a single parent as we are not building a strict hierarchy but rather a directed relational graph where each node represents abstract semantic relations between its descendants.  To illustrate, consider the class \emph{sculpting} from M-MiT which is a child of both \emph{carving}, with \emph{peeling} and \emph{shaving}, and \emph{making art} which includes the descendants \emph{drawing} and \emph{painting} (see Figure \ref{fig:semanticGraph}).  Treating the graph as a hierarchy would be incomplete as we would miss out on the full breadth of relations between these different actions.

\noindent \textbf{Category Embeddings.} To increase the amount of information given to our model in training and to solidify the relationship between an abstraction node and its children, we generate a semantic embedding vector based on the intuition of distributional semantic word representations in natural language processing \cite{NIPS2013_5021,fasttext} for each node. These representations capture contextual word relationships from a large unlabeled corpora.  We use the word embeddings generated by the Subword Information Skip Gram (SISG) model \cite{fasttext} which takes into account the morphology of each word.  These vectorized relationships are complementary to our semantic abstraction graph and aid in allowing our model to capture additional relationships not directly encoded into the graph structure.

\noindent We begin by assigning each leaf node in our graph the average vector of all the words in the class name using SISG.  We then consider this the embedding of that node.  From here we traverse the graph and assign each node an embedding vector that is the average of the embeddings of all of its direct children.  We use this approach for each parent node to ensure that the embeddings of the abstraction nodes are constrained to capture the common relationships among their children while downplaying features that are specific to a single child node.  We describe our approach on using these embeddings to train our model in Section \ref{sec:approach}.

\noindent \textbf{Video Embeddings.} As a first step in training video models to capture the abstract semantic relationships between different classes described above we assign embedding vectors to each video according to their class associations.  For example, a video in the Kinetics dataset with the class \emph{doing aerobics} will be assigned the vector associated with the graph node \emph{doing aerobics}.  This ensures that the models described in Section \ref{sec:approach} learn representations that align with our relational graph.  For a video that contains multiple labels in M-MiT we simply take the average vector for all the classes that belong to the video.

\noindent \textbf{Forming a Training Dataset.} In order to train models to accurately capture abstract semantic relationships among a set of videos we must first build a dataset consisting of sets of videos that share common abstractions.  To do this we iterate through each parent class in our relational graph and sample four videos that belong to any descendant class of the parent.  We use four videos to allow a wide class diversity as using more than four videos in a set commonly results in multiple videos belonging to the same class in both M-MiT and Kinetics.  %
This allows us to balance between having a strong training signal (label diversity) while maximizing computational efficiency.  We then generate a label set such that we find the lowest common abstraction shared between the members of each subset in the power set of the set of the videos.  The abstraction found for a video set of one is simply the label set for that video in the original dataset.  In this way we can train a model to find every abstract semantic relationship present in every combination of the videos in the set greatly increasing our training efficiency over training for different set sizes individually.  The labels generated for each subset are then paired with their associated embedding vectors. Due to the flexible modularity of our architecture we are able to reduce the subsets to only contain pairs when efficiency is a concern.  For this paper we use the full powerset of the input videos to maximize the learning signal for each training step. For Kinetics we generated one million video sets for training and 50k for validation while for M-MiT we generated five million sets for training and 100k for validation.  Training and validation videos were all chosen from the associated training and validation sets of each dataset to preserve the original data splits.

\subsection{Human Performance on Event Abstraction}

We aim to build a model that identifies relationships similar to how humans recognize abstractions between events. First, we collect a human baseline on a video ranking task where the goal is to rank a set of five \emph{query} videos in order of how closely they align with the abstract relationship between a set of \emph{reference} videos.  We compare our trained models to human performance in Section \ref{sec:setCompletion}.

\noindent \textbf{Collecting a Human Baseline Dataset.} Before we collect human baselines we need to build a dataset for ranking videos according to our relational event abstraction paradigm.  We begin similar to the approach used for building our training set and iterate through each abstraction node in our relational graph and select a set of $N$ reference videos (where $N$ can be 1, 2, 3 or 4) that share the abstraction.  From here we calculate a shared embedding vector that is the average between the vectors for each reference video and the vector of their shared abstraction node.  We then sample five query videos from the dataset sorted according to the cosine distance of their embedding vector and this new reference vector.  The goal of this approach is to generate a query set that has at least one video closely aligned with the reference set, one that is very different and three videos that have varying levels of similarity to the reference set.  This forms a range of videos with a quantifiable metric based on reference set similarity which we can use to evaluate human and model performance.

\noindent \textbf{Collecting Human Performance.} To collect human baseline data, we created the Vidrank game (see Supp.) %
and used Amazon Mechanical Turk to crowdsource annotations. Players were presented with a ``Reference'' set of 1-4 videos, and an ``Unknown'' (Query) set of 5 randomly ranked videos labeled from ``Least Similar'' to ``Most Similar'' based on their position. The task was to drag and rearrange the videos in Unknown to a ranking based on each video's similarity to Reference. To ensure reliable results, we required players to pass ``vigilance'' rounds, where it was clear that a video should be placed in the ``Most Similar'' or ``Least Similar'' position. We collected 40 folds of data for each dataset, 10 questions per fold for a total of 800 human responses\footnote{Note that these tasks are challenging for humans who must ignore relations across scenes, colors, etc. The model's advantage is that it is trained only on event relations.}.

\section{Approach} 
\label{sec:approach}

Let $x_i$ be one video, and $\mathcal{X} = \{x_1, \ldots, x_i, \ldots, x_n\}$ be a set of $n$ videos. Given $\mathcal{X}$, our goal is to train a model that correctly predicts the abstract concept that describes all videos in the set and accurately estimates the language model abstraction representation. We will train the model $F(\mathcal{X}; \theta) = (\hat{y}, \hat{e})$ to estimate this category and its semantic word embedding. %
A naive approach is to first classify each individual video in the set with a traditional video classification network (ignoring learned semantic word embeddings), then use these predictions to look up the lowest common ancestor in the graph. However, this approach is problematic. Firstly, the baseline model will be fragile to errors. If the individual classification model makes a mistake, the set abstraction prediction will be wrong. Secondly, since this does not reason about all elements in the set jointly, this baseline will discard useful information, i.e. the abstract category of a video could change depending on the other videos in the set (which our experiments show). Instead, our approach jointly reasons about all elements in the set. 

\begin{figure}[t]
\centering
\includegraphics[width=0.7\linewidth]{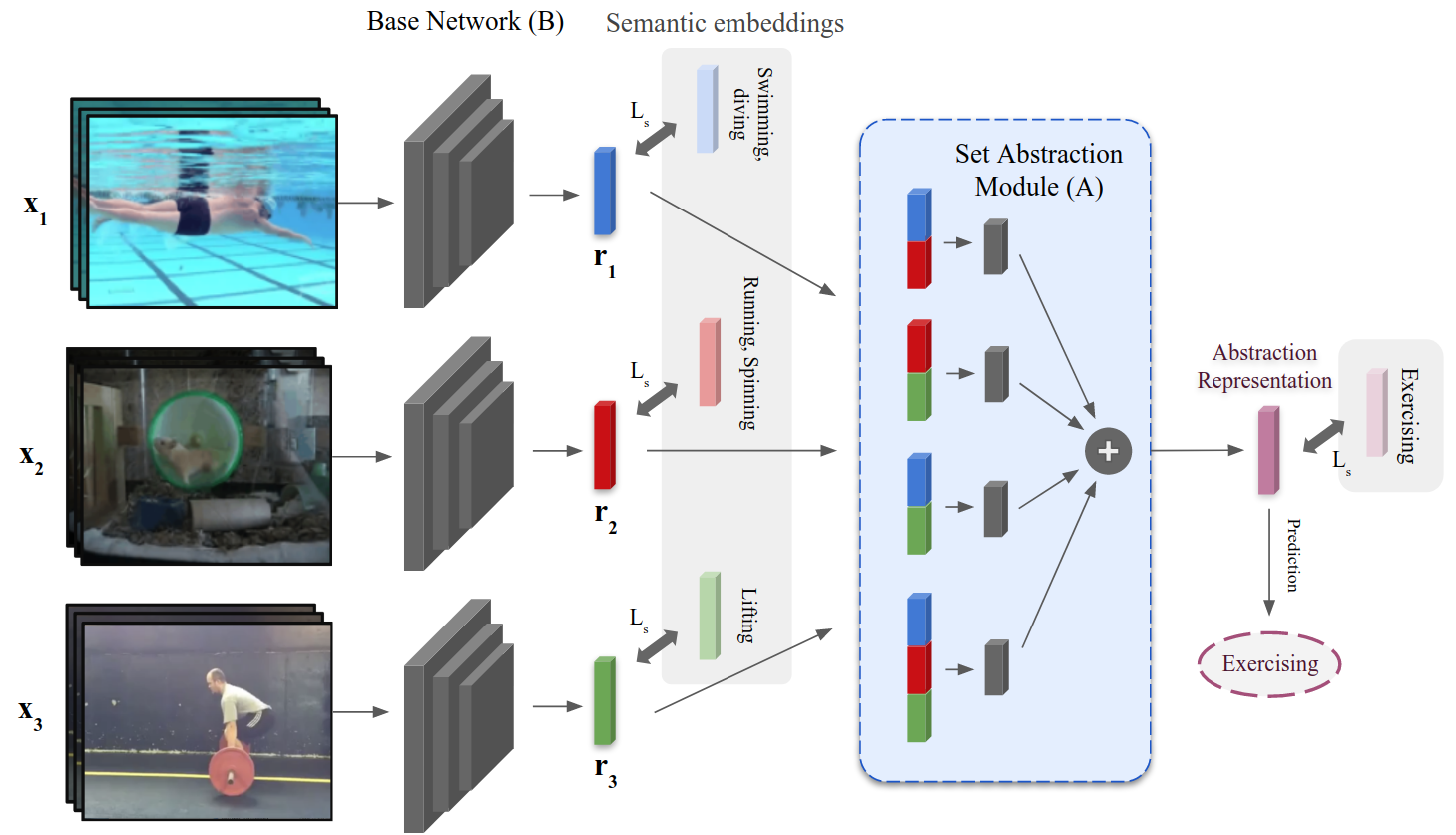}
\caption{\textbf{Set Abstraction Architecture}: A set of $n$ videos ($n=3$ shown) feed into a shared base video model, $B$.  The representations for each video generated by $B$ are combined into each possible subset and fed into the \emph{Set Abstraction Module} (SAM), $A$, which generates a set-wise representation that is used to identify the common abstractions in each subset.  The representations generated by both $A$ and $B$ are contrasted to pretrained semantic embeddings representing the labels for the subset abstractions and the individual videos themselves.}
\label{fig:abstraction_framework}
\end{figure}

\subsection{Relational Set Abstraction Model}

We model $F(\mathcal{X})$ as a deep convolutional network (CNN) that classifies the abstraction of sets of videos. We write $F$ as the composition of two functions $A$ and $B$:
$F(\mathcal{X}) = A\left(\{B(x) | x \in \mathcal{X}\}\right)$
where $A$ is a set abstraction module and $B$ is a base model network for learning individual features. \\

\noindent \textbf{Video Feature Network (\protect{$B$})}
\label{sec:videoNetwork} This network estimates visual features for each individual video in the set. Several base networks have been proposed to handle temporal and motion information in video \cite{carreira2017quo,donahue2015long,simonyan2014two,tran2015learning}. To demonstrate the wide applicability of our framework, we run our experiments with two widely used architectures: a ResNet50 \cite{resNet} with 3D convolutions (ResNet50-3D) and I3Ds \cite{carreira2017quo}. We observed that ResNet50-3D outperformed I3D in most settings, and thus only report ResNet numbers here for clarity (see Supp. for I3D).
Given a set of videos $\mathcal{X} = \{x_0, \ldots, x_n\}$, we use this video feature network to produce a set of features $\mathcal{R} = \{B(x_0), \ldots, B(x_n)\}$, which is fed into the next section. Note that these weights are shared across each video in the set. \\

\noindent \textbf{Set Abstraction Module (\protect{$A$})}
\label{sec:abstractionModule} Given feature embeddings of each video, our set abstraction module (SAM) is trained to predict the common category of all the videos in the set. Rather than learning a representation for one video, the network learns a representation for a set of videos, in sharp contrast with previous works. To correctly recognize abstract categories, the model must capture the relationships between elements in the set. However, there can be multiple relationship orders: within a single video, across videos, and across higher-order tuples. We model all these relationships by operating on the power set, which is the set of all subsets. Our approach will learn features for each subset and also combine them to produce the final abstraction. Specifically, let $g_k(r_1, \ldots, r_k)$ be a neural network that accepts $k$ inputs, and produces features to represent those inputs. This $g$ network will be able to capture the $k$th-order relationship between inputs.
Given a set of features $\mathcal{R}$, the model's prediction can be written as,
{\small $$A(\mathcal{R}) = h\big(\sum_{r_i \in \mathcal{R}} g_1(r_i) + \sum_{r_i,r_j \in \mathcal{R}} g_2(r_i, r_j)
    + \ldots\big).$$}
    
Each $g$ computes features to capture the relationships between its inputs. We exert two sources of supervision onto this representation: (1) we train a shared linear classifier to predict the common abstraction between the inputs of $g$ and (2) train a separate linear layer to estimate the word embedding of the abstraction. The representations produced by each $g_k$  are summed together to create an order invariant representation and the abstract category for the entire set is estimated by another network $h(\cdot)$. Figure \ref{fig:abstraction_framework} illustrates this module.

\subsection{Learning}

The video feature network and the set abstraction module can be trained jointly with stochastic gradient descent. %
As described in Section 5.1, to generate the abstract classification of a set of videos we input the representations computed for each video using the video feature network ($\mathcal{R} = \{B(x_0), \ldots, B(x_n)\}$) into our \emph{set abstraction module} ($g$) and compute the abstract representations for each set of videos ($r_i$) in the power set of input videos ($\mathcal{R}^2$), {\small $A(\mathcal{R}) = g(r_i)$ $\forall$ $r_i \in \mathcal{R}^2.$}

We then apply these representations to two linear models ($h$ and $e$) that capture the abstraction class belonging to the set ($h$) and the category embedding of the abstraction class ($e$) described in Section 3.1.3.  The embedding provides additional supervision and ensures that the representations generated by the model adhere to semantically and contextually relevant features.  

We train the model by averaging the sum of the cross entropy loss for the abstraction prediction, $\mathcal{L}_{ce}$, %
and the mean squared error, $\mathcal{L}_{mse}$, of the generated embedding ($e(g(r_i))$) and the embedding of the ground truth abstraction class ($a_e$) for each subset ($r_i$) in the power set of input videos ($\mathcal{R}^2$): {\small $$\mathcal{L}_{total} = \frac{1}{|\mathcal{R}^2|}\sum_{r_i \in \mathcal{R}^2}\mathcal{L}_{ce}\big(h(g(r_i)), a_c\big) + \mathcal{L}_{mse}\big(e(g(r_i)), a_e\big).$$}

This loss combines the error from generating both the class and the embedding vector of the abstraction class for each possible set of videos given the input set.  By doing this we maximize the supervision signal provided from each video input set without recomputing features for different combinations.  In practice we train with 4 videos producing 15 different video sets (we omit the empty set).

\section{Experiments}

We evaluate our \emph{set abstraction model} on three tasks:
\begin{enumerate}
  \item \textbf{Recognizing set abstractions:} Predict the direct relational abstraction for a set of videos.
  \item \textbf{Set completion:} Given a set of \emph{reference} videos and a set of \emph{query} videos, select the query that best fits in the reference set. 
  \item \textbf{Finding the odd one out:} Identify the video in a set that does not share the relational abstraction common to the other videos in the set.
\end{enumerate}

\noindent \textbf{Experimental Setup.} Our hypothesis is that jointly reasoning about all videos in a set will enable models to more accurately predict set abstractions. We compare our \emph{set abstraction model} (3DResNet50+SAM) against a baseline model that maintains the same base model architecture and is trained for standard classification without a set abstraction module (3DResNet50). As a secondary baseline, we train the same base model for multi-label classification using binary cross entropy loss (3DResNet50+BCE) where the labels consist of all the ancestors of the ground truth class, as well as itself, provided by the dataset.  We use the standard training, validation and test splits provided by M-MiT and Kinetics and show evaluation results for each task on the corresponding test set.\\ 
\noindent \textbf{Comparison to previous work.} To the best of our knowledge, there are no existing video-based models that explicitly address the set abstraction task. Thus, we compare our model to the following extensions of similar previous work:
\begin{itemize}
    \item Relation Networks \cite{santoro2017simple}: we replace SAM with their Relation Module that computes representations over pairs of objects, and use the output of $f_{\phi}$ as the abstraction representation. Importantly, this module can only work with pairs of videos, so we cannot compute results on set completion with N=1.
    \item Odd One Out Networks \cite{Fernando_2017_CVPR}: although their method works on snippets of a single video and only solves the OOO task to generate a representation, we re-purpose their network $(O3N)$ to OOO by extending the number of inputs and training to directly predict the input that does not belong to the set.
\end{itemize}
\noindent \textbf{Implementation Details.} We use PyTorch implementations of 3D ResNet50 \cite{he2016deep} as the basis for our video feature networks. Each $n$-scale relation module in SAM $A$ is a two-layer Multilayer Perceptrons (MLP) with ReLU nonlinearities and 2048 units per layer.  All models were optimized using stochastic gradient descent with a momentum term of 0.9 and weight decay of 5e-4. An initial learning rate of 0.001 was decreased by a factor of 10 every 20 epochs of training. Models were trained until convergence ($\sim 50 - 60$ epochs).

\begin{table}[tb]
\setlength{\tabcolsep}{2.5pt}
\scriptsize
\centering
\begin{tabular}{c | c | c c |  c c | c c }
& & \multicolumn{2}{c|}{\textbf{N = 2}} & \multicolumn{2}{c|}{\textbf{N = 3}} & \multicolumn{2}{c}{\textbf{N = 4}}\\
\textbf{Dataset} &  \textbf{Model} &  \textbf{Top1} & \textbf{Top5} & \textbf{Top1} & \textbf{Top5} & \textbf{Top1} & \textbf{Top5} \\
\hline
\multirow{5}{*}{\textbf{M-MiT}}
& Chance             & 7.9  & 16.2 & 7.9  & 16.2 & 7.9  & 16.2 \\
& 3DResNet50         & 17.1 & 31.2 & 22.6 & 38.8 & 26.0 & 42.9 \\
& 3DResNet50 (BCE)   &  3.9 & 30.0 &  4.9 & 34.5 &  5.1 & 38.0 \\
& 3DResNet50+RN \cite{santoro2017simple}      & 32.4 & 65.2 & 39.2 & 75.4 & 44.9 & 82.1 \\
& 3DResNet50+SAM (Ours)    & \textbf{34.0} & \textbf{66.9} & \textbf{41.1} & \textbf{77.1} & \textbf{47.2} & \textbf{83.8} \\
\hline
\multirow{5}{*}{\textbf{Kinetics}} 
& Chance             & 0.44 & 2.18 & 0.44 & 2.18 & 0.44 & 2.18 \\
& 3DResNet50         & 29.9 & 49.1 & 22.1 & 42.8 & 17.9 & 40.4 \\
& 3DResNet50 (BCE)   &  2.8 & 25.0 &  2.2 & 22.2 &  0.5 & 22.5 \\
& 3DResNet50+RN \cite{santoro2017simple}     & 53.9 & 83.0 & 61.6 & 90.2 & 66.0 & 93.8 \\
& 3DResNet50+SAM (Ours)    & \textbf{60.5} & \textbf{86.0} & \textbf{65.3} & \textbf{91.6} & \textbf{69.9} & \textbf{94.6} \\
\bottomrule
\end{tabular}
\caption{\textbf{Recognizing Set Abstractions:} Classification accuracy (percent) of the models evaluated on the set abstraction task. Here, $N$ is the number of elements in the set, and the top$k$ chance level is the sum of the frequency of the top$k$ most frequent abstract nodes presented during evaluation.}
\label{table:abstraction_result}
\end{table}

\subsection{Recognizing Set Abstractions}

\begin{figure}[t]
\centering
\includegraphics[width=0.85\linewidth]{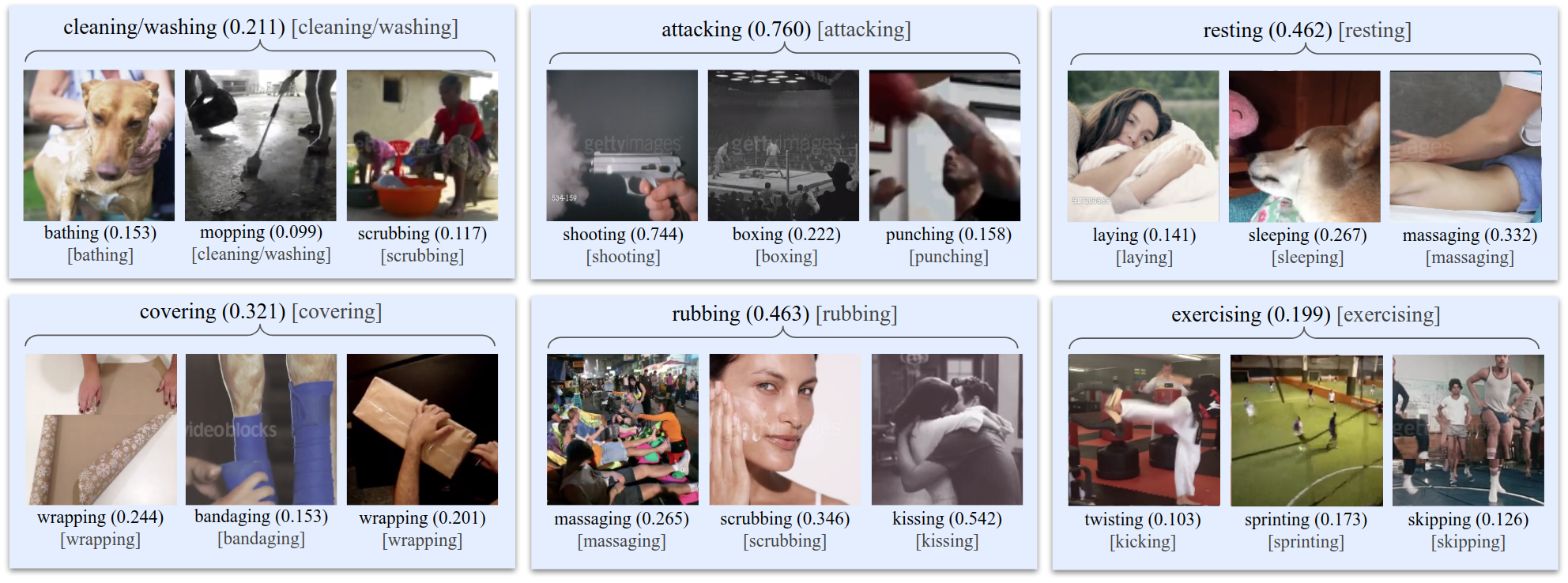}
\caption{\textbf{Qualitative Set Abstraction Results.} We show results for sets of length 3, and only show the predictions for the individual videos and the entire set to simplify the visualization. Confidence is indicated in parenthesis and ground truth class in brackets.}
\label{fig:set_abs_results}
\end{figure}

We first evaluate our model on recognizing the abstract category of a set of $N$ videos. We use our abstraction model to directly predict this category given the set.  Our baseline model (3DResNet50) individually predicts the specific class category for each video in the set, then computes the set abstraction class directly from the semantic graph (Section \ref{sec:dataset}) based on its predictions.  Our multi-label baseline (3DResNet50+BCE) also evaluates each video independently and selects the abstract category with the highest mean probability across videos. Our results suggest that there are significant gains by jointly modeling all elements in the video set. Table \ref{table:abstraction_result} quantitatively compares the proposed \emph{set abstraction model} against the baselines and Fig. \ref{fig:set_abs_results} shows qualitative results. Since the margin of improvement increases with the size of the input set, this suggests that set-based training improves the strength of the learning signal by reducing ambiguity.

\subsection{Set Completion}
\label{sec:setCompletion}
\begin{figure}[t]
\centering
\includegraphics[width=0.85\linewidth]{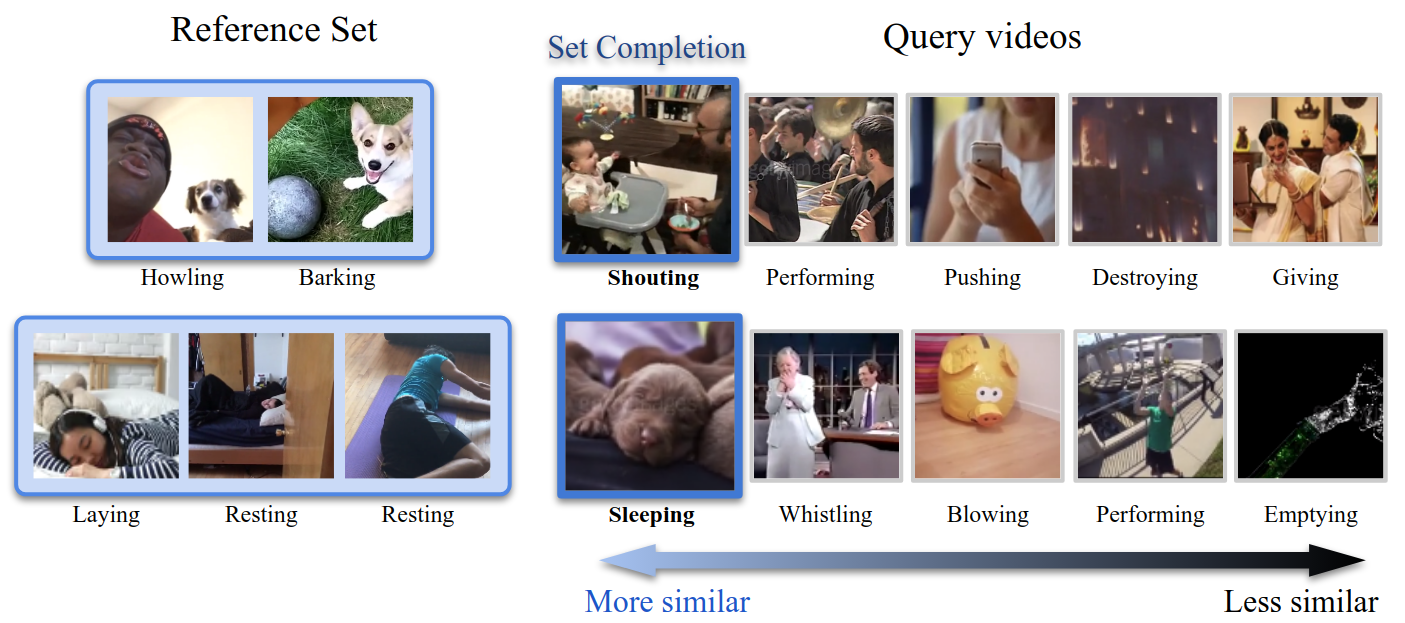}
\caption{\textbf{Set Completion via Learned Abstractions.} The reference videos are on the left, and the query videos show top ranked retrievals that complete them. The labels of individual videos (not provided to the model) are listed below. The model is able to understand the underlying abstraction regardless of the subject, e.g. choosing the sleeping dog despite all human subjects in the reference.
\label{fig:abstraction_query}}
\end{figure}

While recognizing the abstract event relationships shared among a set of videos shows that our model has learned to identify common patterns, we aim to solve more cognitive-level tasks.  Thus, we apply our model to the complex task of ranking a set of five \emph{query} videos according to how closely they align to the common abstraction found in a set of $N\in[1, 4]$ \emph{reference} videos (see Figure \ref{fig:abstraction_query}).

First we use the model to generate an abstract representation of the \emph{reference} set of videos using the method from Section \ref{sec:abstractionModule}.  Then we rank each \emph{query} video according to the cosine distance between this abstract representation and their single video feature representations (Section \ref{sec:videoNetwork}). 
This distance tells how closely each video is aligned to the \emph{reference} set abstraction found by our model.

\begin{table}[tb]
\setlength{\tabcolsep}{5pt}
\centering
\scriptsize
\begin{tabular}{c | c | c | c | c | c | c}
\toprule
\textbf{Dataset} & \textbf{Model} & \textbf{N = 1} & \textbf{N = 2} & \textbf{N = 3} & \textbf{N = 4} & \textbf{Avg} \\
\hline
\multirow{4}{*}{\textbf{M-MiT}}
& Human Baseline        & 0.547 & 0.495          & 0.595 & 0.541          & 0.545 \\
\hline
& 3DResNet50            & 0.388          & 0.415          & 0.455          & 0.463          & 0.430 \\
& 3DResNet50+RN \cite{santoro2017simple}        & -              & 0.483          & 0.513          & 0.533          & 0.489 \\
& 3DResNet50+SAM (Ours)      & \textbf{0.481}          & \textbf{0.544} & \textbf{0.570}  & \textbf{0.571} & \textbf{0.542}  \\
\hline
\multirow{4}{*}{\textbf{Kinetics}}
& Human Baseline         & 0.432          & 0.653 & 0.629          & 0.606 & 0.58
\\
\hline
& 3DResNet50             & 0.339          & 0.421          & 0.431          & 0.459          & 0.413 \\
& 3DResNet50+RN \cite{santoro2017simple}         & -              & 0.491          & 0.487          & 0.489          & 0.489 \\
& 3DResNet50+SAM (Ours)         & \textbf{0.523} & \textbf{0.627}   & \textbf{0.659} & \textbf{0.606} & \textbf{0.604} \\
\bottomrule
\end{tabular}
\caption{\textbf{Set Completion:} Rank Correlation of our model (3DResNet50+SAM), a baseline (3DResNet50) and human ranking to the ranking achieved using the embedding distance between a video and the abstraction of a \emph{reference} set of size $N$ on the \emph{set completion} task.}
\label{table:abstraction_retrieval}
\end{table}

To evaluate our results, we correlate our model's ranking order with the ground truth order found by using the natural-language embedding distance (Section \ref{sec:dataset}).  In this way we compare the performance to both the abstract relationships defined in our \emph{Semantic Relational Graph} and event embeddings associated with the annotated labels of each video.  We evaluate the results with, and without, the proposed \emph{abstraction module} (3DResNet50+SAM and 3DResNet50 respectively).  For the model without the \emph{abstraction module} we rank the \emph{query} videos using the cosine distance to the average of the feature vectors from each individual \emph{reference} video. Since we are interested in developing models for human-level understanding of abstract relationships, we additionally compare our model results to human performance on the same video ranking task (see Section \ref{sec:dataset}). Table \ref{table:abstraction_retrieval} summarizes our results.  Our abstraction model (3DResNet50+SAM) beats the human baseline on M-MiT when the \emph{reference} set has either 2 or 4 videos and only underperforms the human baseline on Kinetics when the \emph{reference} set has 3 videos. We can see that our model achieves near human performance on M-MIT and surpasses human performance on Kinetics.

\subsection{Finding The Odd One Out}

Given a set of videos, which does not belong? With a model trained for set abstraction, we can use its learned representation for new tasks, such as identifying the odd one out in a set, without additional training. For example, given a set of events (\emph{barking}, \emph{boating} and \emph{flying}), the correct outlier would be \emph{barking} since \emph{boating} and \emph{flying} are both instances of \emph{traveling}, while \emph{barking} is not.

\begin{figure}[t]
\centering
\includegraphics[width=0.75\linewidth]{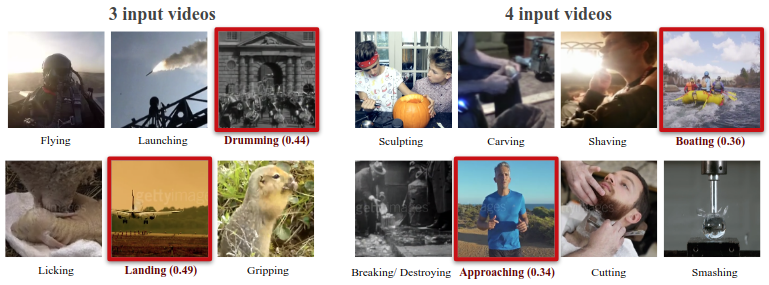}
\caption{\textbf{Qualitative Results for Odd One Out detection}: Given sets of three videos (left) and four videos (right), which one is odd? The odd video detected by our \emph{set abstraction model} is indicated by a red bounding box (probability in parenthesis). Even with a small number of other videos to compare to, the model is able to select the odd video out.}
\label{fig:outlier_qualitative}
\end{figure}

\begin{table}[tb]
\setlength{\tabcolsep}{5pt}
 \scriptsize
\centering
\begin{tabular}{c | c | c | c | c | c}
\toprule
& & \multicolumn{2}{c|}{\textbf{N = 3}} & \multicolumn{2}{c}{\textbf{N = 4}} \\
\textbf{Dataset} & \textbf{Model} &  \textbf{Top-1} & \textbf{Top-2} & \textbf{Top-1} & \textbf{Top-2} \\
\hline
\multirow{4}{*}{\textbf{M-MiT}} &
Human Baseline           & 74.21 & - & 78.04 & - \\
\hline
& 3DResNet50       & 49.95 & 78.02 & 43.73 & 68.34 \\
& 3DResNet50+RN  \cite{santoro2017simple}  & 36.20 & 64.30 & 28.94 & 50.71 \\
& 3DResNet50+O3N \cite{Fernando_2017_CVPR}  & 34.10 & 60.11 & 35.84 & 60.71 \\
& 3DResNet50+SAM (Ours)  & \textbf{52.63} & \textbf{79.86} & \textbf{47.21} & \textbf{71.11} \\
\hline
\multirow{4}{*}{\textbf{Kinetics}} &
Human Baseline     & 87.31 & - & 85.40 & - \\
\hline
& 3DResNet50     & 65.15 & 82.65 & 69.62 & 81.48 \\
& 3DResNet50+RN \cite{santoro2017simple}  & 40.11 & 70.97 & 30.48 & 54.41 \\
& 3DResNet50+O3N \cite{Fernando_2017_CVPR} & 55.14 & 80.59 & 66.00 & 81.80 \\
& 3DResNet50+SAM (Ours) & \textbf{85.90} & \textbf{92.80} & \textbf{83.18} & \textbf{91.44} \\
\bottomrule
\end{tabular}
\caption{\textbf{Odd One Out detection accuracy:} Predict the element that does not belong to the set. The language-enhanced features from the set abstraction network are compared with the features from the corresponding base model.}
\label{table:baselines_outlier}
\end{table}

Given a set of videos, we identify the odd one out by choosing the video with the largest cosine distance between its video representation and the \emph{abstract} representation of the remaining videos in the set (Section \ref{sec:abstractionModule}).
Intuitively, this method leverages a model’s ability to preserve ``conceptual'' distance in its feature space.
Table \ref{table:baselines_outlier} shows that our \emph{set abstraction model} achieves good performance without additional training by making pairwise distance comparisons on subsets of the input. The abstraction model consistently outperformed the baseline model suggesting that our approach indeed learns stronger set representations.   We show qualitative examples of our performance in Figure \ref{fig:outlier_qualitative}.

\section{Conclusion}
A central challenge in computer vision is to learn abstractions of dynamic events. By training models to capture semantic relationships across diverse events and predict common patterns, we show that we can learn rich representations of similarity for a new set of tasks. By rooting our models in language, the model can learn abstractions that are better suited to represent how people form high-level classes. %
Recognizing abstractions should enable vision systems to summarize high-level patterns for different types of applications. 
While our focus is on capturing action relationships, future work could take into account abstractions involving scenes, objects and other concepts to provide a larger range of relationships to understand events (e.g. ``driving'' and ``jogging'' may both occur on a ``road'' while ``writing'' and ``drawing'' may both use a ``pencil'').

\noindent \textbf{Acknowledgments.} This work was supported by the MIT-IBM Watson AI Lab (to R.F and A.O), NSF IIS 1850069 (to C.V) and the Intelligence Advanced Research Projects Activity (IARPA) via Department of Interior/ Interior Business Center (DOI/IBC) contract number D17PC00341. The U.S. Government is authorized to reproduce and distribute reprints for Governmental purposes notwithstanding any copyright annotation thereon.

\noindent \textbf{Disclamer.} The views and conclusions contained herein are those of the authors and should not be interpreted as necessarily representing the official policies or endorsements, either expressed or implied, of IARPA, DOI/IBC, or the U.S. Government.

\clearpage
\bibliographystyle{splncs04}
\bibliography{egbib}
\end{document}